\documentclass{article}

\usepackage[preprint]{neurips_2026}

\usepackage[utf8]{inputenc}
\usepackage[T1]{fontenc}
\usepackage{hyperref}
\usepackage{url}
\usepackage{booktabs}
\usepackage{amsfonts}
\usepackage{nicefrac}
\usepackage{microtype}
\usepackage{graphicx}
\usepackage{amsmath,amssymb}
\usepackage{dsfont}

\title{Perturbation Sensitivity at Convergence: A Simple Signal for Identifying Spuriously Correlated Samples}

\author{
  Nilesh Kumar \\
  \texttt{nk4856@rit.edu} \\
}

\begin{document}
\maketitle

\begin{abstract}

Models trained by empirical risk minimization on data containing spurious correlations achieve high average accuracy while failing on subpopulations where the correlation does not hold. Existing methods for identifying the affected samples without group annotations rely on signals from early training, which requires locating the epoch at which to intervene, a hyperparameter typically selected using group-labeled validation data. We show that a usable signal is available after convergence, when loss no longer distinguishes the two populations. Samples consistent with the spurious correlation are classified by a shared rule, while the remaining samples are fit through configurations specific to individual inputs and are correspondingly more fragile. Applying a fixed perturbation to a converged model's inputs flips the predictions of the latter far more often than the former. The resulting procedure requires two forward passes per training sample, no group annotations at any stage, and no early-stopping epoch. Using the detected samples to rebalance training raises worst-group accuracy on Waterbirds from 57.3\% to 80.8\%, against 85.8\% with ground-truth group labels.

\end{abstract}

\section{Introduction}

Neural networks trained by empirical risk minimization on real-world data frequently learn feature-label associations that hold in the training distribution but do not reflect the underlying task. The canonical illustration is a classifier that separates cows from camels by background, grass versus sand, rather than by the animal itself. Such a model attains high average accuracy while failing on the subpopulations where the correlation is absent: cows on sand, camels on grass. This gap between average and worst-group accuracy is a persistent obstacle to deploying models in settings where subgroup reliability matters.

A substantial body of work addresses this by identifying the samples on which the spurious rule does not apply, the non-SC or minority samples, and upweighting them during a second round of training. The difficulty is identification. Group annotations are expensive and, in realistic settings, unavailable; methods that infer groups without them must find some proxy signal in the training dynamics.

Existing proxies are, with few exceptions, drawn from early training. JTT \cite{liu2021just} retrains on the samples an intermediate model misclassifies; SPARE \cite{yang2024identifying} clusters model outputs in the first one or two epochs. The shared premise is that SC and non-SC samples are distinguishable during a transient window and not afterwards, because an overparameterized network eventually drives training loss to near zero on both. This premise is well founded but has a cost: it makes the methods dependent on locating the window. The epoch at which to intervene becomes a hyperparameter, and in practice it is selected using group-labeled validation data, reintroducing the supervision the methods were designed to avoid.

We make a different observation. Convergence does not erase the distinction between SC and non-SC samples; it changes which measurement reveals it. SC samples are classified by a simple rule shared across a large subpopulation, and that rule is not specific to any individual input. Non-SC samples cannot be classified by that rule, and the network fits them instead through configurations tailored to individual samples, memorization. Fitting a sample individually produces a more fragile solution than fitting it by a shared rule, and fragility is directly measurable: perturb the input and see whether the prediction survives.

This yields a detection procedure of unusual simplicity. Take a converged model, apply a fixed perturbation to each training image, and record whether the predicted label changes. Samples whose predictions flip are the memorized ones. There is no early-stopping epoch to locate, no continuous score to threshold, no group annotations at any stage, and no retraining of the detector. The procedure requires two forward passes per training sample and applies to any model already trained.

We find that this signal separates SC from non-SC samples on Waterbirds \cite{sagawa2019distributionally}, and that using it to rebalance training recovers most of the benefit of ground-truth group labels. On Waterbirds, worst-group accuracy rises from 57.3\% under ERM to 80.8\%, against 85.8\% when true group labels are used. In preliminary experiments, the separation appears specific to convergence, degrading when measured at earlier checkpoints.

This report documents the observation and the supporting experiments. It is not a complete empirical study: We aim to establish that a signal this simple exists and works.

\section{Method}

We assume a model trained to convergence by standard empirical risk
minimization, with no group annotations available at any stage.

\paragraph{Detection.}\label{sec:detection}
Let $f$ denote the converged model and $T$ a fixed input perturbation.
For each training sample $x_i$ we compute a binary sensitivity indicator
\begin{equation}
s_i = \mathds{1}\!\left[\arg\max f(T(x_i)) \neq \arg\max f(x_i)\right],
\end{equation}
which records whether the perturbation changes the predicted class.
We use Gaussian blur as the perturbation, with kernel size $k=21$ for
Waterbirds. The kernel is set to span roughly a
tenth of the image width, large enough to destroy fine-scale detail while
preserving coarse structure and color.

Note that $s_i$ is threshold-free: the prediction either changes or it does
not, so no cutoff on a continuous score is required.

\paragraph{Pseudo-groups.}
Samples are partitioned within each class $c$ into two pseudo-groups indexed by
\begin{equation}
g_i = 2c + s_i,
\end{equation}
giving $2|\mathcal{C}|$ pseudo-groups in total. Partitioning within rather
than across classes matches the group structure of the benchmarks, in which
the spurious attribute is defined relative to the class label.

\paragraph{Rebalancing.}\label{sec:rebalancing}
We retrain from the same initialization used for the first stage, an
ImageNet-pretrained ResNet-50 for Waterbirds using a weighted sampler in which each pseudo-group contributes
equally in expectation. Sample $i$ is assigned weight $w_i = 1/|G_{g_i}|$,
where $G_g$ denotes the set of samples in pseudo-group $g$. No other change
is made to the training procedure.

\begin{table}[t]
\centering
\caption{Fraction of samples flagged by perturbation sensitivity ($k=21$),
broken down by ground-truth group. Detection uses no group labels; groups
are shown only for evaluation. Minority groups are flagged at roughly ten
times the rate of majority groups.}
\label{tab:detection}
\begin{tabular}{llrr}
\toprule
Group & Type & $n$ & Flagged \\
\midrule
$g_0$ landbird on land   & majority & 3498 & 7.3\% \\
$g_1$ landbird on water  & minority & 184  & 70.7\% \\
$g_2$ waterbird on land  & minority & 56   & 69.6\% \\
$g_3$ waterbird on water & majority & 1057 & 3.5\% \\
\midrule
\multicolumn{2}{l}{Minority recall} & & 70.4\% \\
\multicolumn{2}{l}{Minority precision} & & 36.6\% \\
\bottomrule
\end{tabular}
\end{table}

\subsection{Perturbation sensitivity identifies minority samples}

Table~\ref{tab:detection} reports the fraction of Waterbirds training samples
flagged by the detection procedure, broken down by ground-truth group. Group
labels are used only for this evaluation and play no part in the procedure
itself.

The separation is substantial. Samples in the two minority groups are flagged
at roughly ten times the rate of samples in the two majority groups, 70.7\%
and 69.6\% against 7.3\% and 3.5\%. The effect is consistent across both
minority groups despite a large difference in their sizes, 184 samples for
landbirds on water and 56 for waterbirds on land, indicating that the signal
does not depend on a group being large enough to influence the loss landscape
collectively.

Detection is imperfect in a specific and, for our purposes, benign way. Recall
over the two minority groups is 70.4\%, while precision is 36.6\%: the
procedure flags 462 samples in total, of which 169 belong to a true minority
group. The remaining 293 are majority samples that the converged model
nonetheless classifies fragilely. The consequence of a false positive is that a
majority sample is upweighted during the second training stage, which is far
less damaging than the converse error of leaving a minority sample
downweighted. Section~\ref{sec:rebalancing} shows that the rebalancing result
is robust to this asymmetry.

\begin{table}[t]
\centering
\caption{Waterbirds test accuracy. \textsc{Up-ps} rebalances using
pseudo-groups from perturbation sensitivity; \textsc{Up-GT} uses ground-truth
group labels. All three share architecture, initialization, optimizer, and
epoch budget, differing only in the sampling distribution.}
\label{tab:waterbirds}
\begin{tabular}{lrrr}
\toprule
& ERM & \textsc{Up-ps} & \textsc{Up-GT} \\
\midrule
Overall accuracy      & 85.8 & 88.9 & 92.3 \\
Worst-group accuracy  & 57.3 & 80.8 & 85.8 \\
\midrule
$g_0$ landbird on land   & 99.7 & 97.8 & 96.8 \\
$g_1$ landbird on water  & 77.8 & 81.0 & 89.7 \\
$g_2$ waterbird on land  & 57.3 & 80.8 & 85.8 \\
$g_3$ waterbird on water & 93.6 & 93.5 & 91.9 \\
\bottomrule
\end{tabular}
\end{table}

\subsection{Rebalancing on detected samples}
\label{sec:rebalancing}

Table~\ref{tab:waterbirds} reports test accuracy after retraining with the
pseudo-groups of Section~\ref{sec:detection}. Worst-group accuracy rises from
57.3\% under ERM to 80.8\%. The improvement is concentrated in the two
minority groups, with the majority groups losing between one and two points,
the expected consequence of redistributing sampling weight away from them.

We note that the detection stage over-flags by roughly a factor of two
relative to the true minority counts, yet the downstream gain is largely
preserved. This is consistent with the asymmetry described above: precision
matters less than recall for this use, because the cost of upweighting a
correctly-classified majority sample is small.

\section{Limitations}

This report documents a single observation with the minimum evidence needed
to establish it, and several things are absent. We evaluate on one dataset,
Waterbirds, and report single-seed results. We do not compare against
published group-inference methods under matched conditions. We have not
swept the perturbation kernel size on this pipeline, so the robustness of the
result to that choice is untested. We also do not systematically report how the signal
varies with the checkpoint at which it is measured; preliminary evidence
suggests the separation is specific to convergence and degrades at earlier
checkpoints, which would be consistent with the account given above, but we
defer this to future work.

\bibliographystyle{plainnat}
\bibliography{ref}

\end{document}